\documentclass[10pt,twocolumn,letterpaper]{article}

\usepackage{cvpr}
\usepackage{times}
\usepackage{epsfig}
\usepackage{graphicx}
\graphicspath{{./}{./images/}}

\usepackage{amsmath}
\usepackage{amssymb}
\usepackage{mystyle}

\usepackage{subcaption}

\usepackage[pagebackref=true,breaklinks=true,letterpaper=true,colorlinks,bookmarks=false]{hyperref}

\cvprfinalcopy 

\def\cvprPaperID{4208}

\ifcvprfinal\fi

\begin{document}

\title{GAGAN: Geometry-Aware Generative Adversarial Networks}

\author{Jean Kossaifi\(^{*}\) \qquad
 Linh Tran\(^{*}\) \qquad
 Yannis Panagakis\(^{*, \dagger}\) \qquad
 Maja Pantic\(^{*}\)\\
\(^{*}\)Imperial College London\\
\(^{\dagger}\) Middlesex University London\\
{\tt\small \{jean.kossaifi;linh.tran;i.panagakis;m.pantic\}@imperial.ac.uk}
}

\maketitle

\begin{abstract}
Deep generative models learned through adversarial training have become increasingly popular for their ability to generate naturalistic image textures.  However, aside from their texture, the visual appearance of objects is significantly influenced by their shape geometry; information which is not taken into account by existing generative models.
This paper introduces the Geometry-Aware Generative Adversarial Networks (GAGAN) for incorporating geometric information into the image generation process. Specifically, in GAGAN the generator samples latent variables from the probability space of a statistical shape model. 
By mapping the output of the generator to a canonical coordinate frame through a differentiable geometric transformation, we enforce the geometry of the objects and add an implicit connection from the prior to the generated object. 
Experimental results on face generation indicate that the GAGAN can generate realistic images of faces with arbitrary facial attributes such as facial expression, pose, and morphology, that are of better quality than current GAN-based methods. 
Our method can be used to augment any existing GAN architecture and improve the quality of the images generated.
\end{abstract}

\section{Introduction}
\label{sec:introduction}
\begin{figure}
\includegraphics[width=\linewidth]{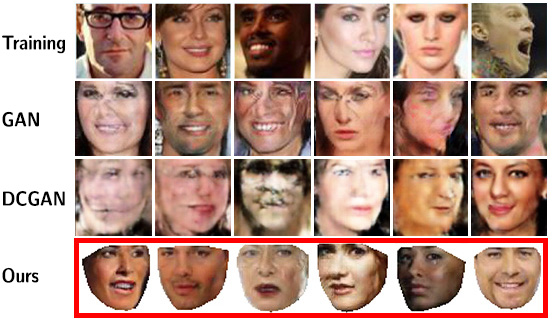}
\caption{\textbf{Samples generated by GANs trained on the CelebA \cite{liu2015faceattributes}}. The first row shows some real images used for training. The middles rows present results obtained with popular GAN architectures, namely DCGAN \cite{radford2015unsupervised} (row 2) and WGAN \cite{arjovsky2017wasserstein} (row 3). 
Images generated by our proposed GAGAN architecture (last row) look more realistic and the represented objects follows an imposed geometry, expressed by a given shape prior.}
\label{fig:intro_bad_examples}
\end{figure}

Generating images that look authentic to human observers is a longstanding problem in computer vision and graphics.
Benefitting from the rapid development of deep learning methods and the easy access to large amounts of data, image generation techniques have made significant advances in recent years. 
In particular, Generative Adversarial Networks \cite{goodfellow2014generative} (GANs) have become increasingly popular for their ability to generate visually pleasing results, without the need to explicitly compute probability densities over the underlying distribution.

However, GAN-based models still face many unsolved difficulties. 
The visual appearance of objects is not only dictated by their visual texture but also depends heavily on their shape geometry. 
Unfortunately, GANs do not allow to incorporate such geometric information into the image generation process. As a result, the shape of the generated visual object cannot be explicitly controlled. This significantly degenerates the visual quality of the produced images. Figure~\ref{fig:intro_bad_examples} demonstrates the challenges for face generation with different GAN architectures (DCGAN \cite{radford2015unsupervised} and WGAN \cite{arjovsky2017wasserstein}) that have been trained on the celebA dataset \cite{liu2015faceattributes}. Whilst GANs \cite{goodfellow2014generative, radford2015unsupervised} and Wasserstein GANs (WGANs) \cite{arjovsky2017wasserstein} generate crisp realistic objects (e.g. faces), their geometry is not followed. 
There have been attempts to include such information in the prior, for instance the recently proposed Boundary Equilibrium GANs (BEGAN) \cite{berthelot2017began}, or to learn latent codes for identities and observations \cite{donahue2017semantically}. However, whilst these approaches in some cases improved image generation, they still fail to explicitly model the geometry of the problem.
As a result, the wealth of existing annotations for fiducial points, for example from the facial alignment field, as well as the methods to automatically and reliably detect those \cite{bulat2017far}, remain largely unused in the GAN literature.

In this paper, we address the challenge of incorporating geometric information about the objects into the image generation process.
To this end, the Geometry-Aware GAN (GAGAN) is proposed in Section \ref{sec:method}. Specifically, in GAGAN the generator samples latent variables from the probability space of a statistical shape model. By mapping the output of the generator to the coordinate frame of the mean shape through a differentiable geometric transformation, we implicitly enforce the geometry of the objects and add an implicit skip connection from the prior to the generated object. The proposed method exhibits several advantages over the available GAN-based generative models, allowing the following contributions:
\begin{itemize}
\item GAGAN can be easily incorporated into and improve any existing GAN architecture
\item GAGAN generates morphologically-credible images using prior knowledge from the data distribution (adversarial training) and allows to control the geometry of the generated images
\item GAGAN leverages domain specific information such as symmetry and local invariance in the geometry of the objects as additional prior. This allows to exactly recover the lost information inherent in generation from a small latent space 
\item By leveraging the structure in the problem, unlike existing approaches, GAGAN works with small datasets (less than \(25,000\) images). 
\end{itemize}

We assessed the performance of GAGAN in Section~\ref{sec:experiment} by conducting experiments on face generation. The experimental results indicate that GAGAN produces superior results with respect to the visual quality of the images produced by existing state-of-the-art GAN-based methods. In addition, by sampling from the statistical shape model we can generate faces with arbitrary facial attributes such as facial expression, pose, and morphology.
 
\section{Background and related work}
\label{sec:background}
\begin{figure*}[ht!]
\begin{center}
   \includegraphics[width=1\linewidth]{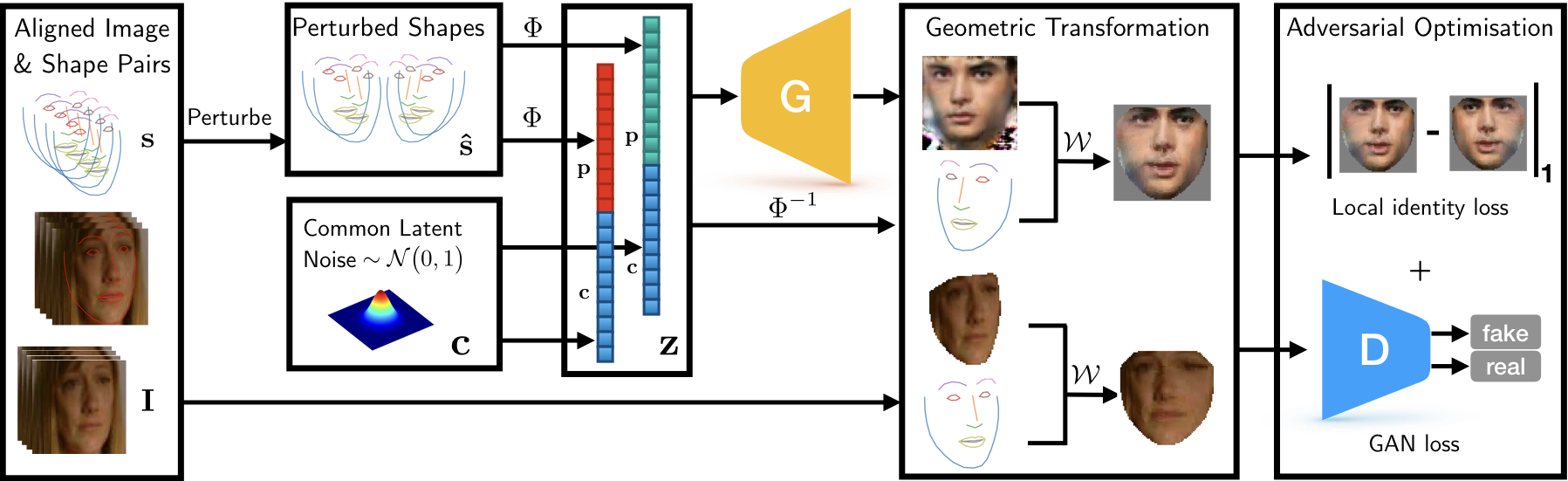}
\end{center}
   \caption{\textbf{Overview of our proposed GAGAN method}.
   (i) For each training image \(\mymatrix{I}\), we leverage the corresponding shape~\(\myvector{s}\). Using the geometry of the object, as learned in the statistical shape model, perturbations \( \myvector{\hat{s}}_1, \cdots,  \myvector{\hat{s}}_n\) of that shape are created.
   (ii) These perturbed shapes are projected onto a normally distributed latent subspace using the normalised statistical shape model. That projection \(\Phi\bm(\myvector{s}\bm)\) is concatenated with a latent component \(\myvector{c}\), shared by all perturbed versions of a same shape.
   (iii) The resulting vectors \(\myvector{\hat z}_1, \cdots, \myvector{\hat z}_n\) are used as inputs to the Generator which generate fake images \(\mymatrix{\hat I}_1, \cdots, \mymatrix{\hat I}_n\). The geometry imposed by the shape prior is enforced by a geometric transformation \(\mathcal{W}\) (in this paper, a piecewise affine warping) that, given a shape \(\myvector{\hat s}_k\), maps the corresponding image \(\mymatrix{\hat I}_k\) onto the canonical shape. These images, thus normalised according to the shape prior, are classified by the Discriminator as fake or real. The final loss is the sum of the GAN loss and an \(\ell_1\) loss enforcing that images generated by perturbations of the same shape be visually similar in the canonical coordinate frame.
 }
\label{fig:GAGAN}
\end{figure*}

\paragraph{Generative Adversarial Networks} \hspace{-0.75em} \cite{goodfellow2014generative} approach the training of deep generative models from a game theory perspective using a minimax game. That is, GANs learn a distribution $P_G(\myvector{x})$ that matches the real data distribution $P_{data}(\myvector{x})$, hence their ability to generate new image instances by sampling from $P_G(\myvector{x})$.  Instead of explicitly assigning a probability to each point in the data distribution, the generator G learns a (non-linear) mapping function from a prior noise distribution $P_\myvector{z}(\myvector{z})$ to the data space as $G(\myvector{z}; \theta)$. This is achieved during training, where the generator G ``plays'' a zero-sum game against an adversarial discriminator network D. The latter aims at distinguishing between fake samples from the generator's distribution $P_G(\myvector{x})$ and real samples from the true data distribution $P_{data}(\myvector{x})$. For a given generator, the optimal discriminator is then 
$D(\myvector{x}) = \frac{P_{data}(\myvector{x})}{P_{data}(\myvector{x}) + P_G(\myvector{x})}$. 
Formally, the minimax game is:
\begin{align}
\begin{split}\nonumber
\min_G \max_D V(D, G) = &
\myexpectation{\myvector{x} \sim P_{data}}{\log D(\myvector{x})}\, + \\
& \myexpectation{\myvector{z} \sim noise}{\log\mylp 1 - D\bm(G(\myvector{z})\bm)\myrp}
\end{split}
\end{align}

The ability to train extremely flexible generating functions, without explicitly computing likelihoods or performing inference, while targeting more mode-seeking divergences, has made GANs extremely successful in image generation  \cite{radford2015unsupervised, odena2016conditional,mirza2014conditional,salimans2016improved}. The flexibility of GANs has also enabled various extensions, for instance to support structured prediction \cite{mirza2014conditional,odena2016conditional}, to train energy based models \cite{zhao2016energy} and combine adversarial loss with an information loss \cite{chen2016infogan}. Additionally, GAN-based generative models have found numerous applications in computer vision, including text-to-image \cite{reed2016generative,zhang2016stackgan}, image-to-image\cite{zhu2017unpaired,isola2016image}, style transfer \cite{johnson2016perceptual}, image super-resolution \cite{ledig2016photo} and image inpainting \cite{pathak2016context}.

However, most GAN formulations employ a simple input noise vector \(\myvector{z}\) without any restriction on the manner in which the generator may use this noise. As a consequence, it is impossible for the latter to disentangle the noise and \(\myvector{z}\) does not correspond to any semantic feature of the data. However, many domains naturally decompose into a set of semantically meaningful latent representations. For instance, when generating faces for the celebA dataset, it would be ideal if the model automatically chose to allocate continuous random variables to represent different factors, e.g. head pose, expression and texture. This limitation is partially addressed by recent methods \cite{chen2016infogan, mathieu2016disentangling,wang2017tag,tran2017disentangled,donahue2017semantically} that are able to learn meaningful latent spaces, explaining generative factors of variation in the data. However, to the best of our knowledge, there has been no work explicitly disentangling the latent space for object geometry of GANs.

\paragraph{Statistical Shape Models} \hspace{-0.75em} were first introduced by Cootes et al. in \cite{cootes1995active} where the authors argue that existing methods tend to favor variability over simplicity and, in doing so, sacrifice model specificity and robustness during testing.
The authors propose to remedy this by building a statistical model of the shape able to deform only to represent the object to be modeled, in a way consistent with the training samples.
This model was subsequently improved upon with Active Appearance Models (AAMs) to not only model the shape of the objects but also their textures \cite{cootes1998interpreting,cootes2001active}.
AAMs operate by first building a statistical model of shape. All calculations are then done in a shape variation-free canonical coordinate frame.
The texture in that coordinate frame is expressed as a linear model of appearance.
However, using row pixels as features for building the appearance model does not yield satisfactory results.
Generally, the crux of successfully training such a model lies in constructing an appearance model rich and robust enough to model the variability in the data. 
In particular, as is the case in most applications in computer vision, changes in illumination, pose and occlusion are particularly challenging.
There has been extensive efforts in the field to design features robust to these changes such as Histograms of Oriented Gradients \cite{dalal2005histograms} (HOG), Image Gradient Orientation kernel (IGO) \cite{tzimiropoulos2012subspace}, Local Binary Patterns \cite{lbp} (LBP) or SIFT features \cite{SIFT}. The latter are considered the most robust for fitting AAMs \cite{antonakos2015feature}.
Using these features, AAMs have been shown to give state-of-the-art results in facial landmarks localisation when trained on data collected in-the-wild \cite{tzimiropoulos2016fast,GN-DPM,antonakos2015feature,kossaifi2017fast,tzimiropoulos2017fast}. Their generative nature make them more interpretable than discriminative approaches while they require less data than deep approaches.
Lately, thanks to the democratisation of large corpora of annotated data, deep methods tend to outperform traditional approaches for areas such as facial landmarks localisation, including AAMs, and allow learning the features end-to-end rather than relying on hand-crafted ones. However, the statistical shape model employed by Active Appearance Model has several advantages. By constraining the search space, the statistical shape model allows methods that leverage it to be trained with smaller dataset. Generative by nature, it is also interpretable and as such can be used to sample new sets of points, unseen during training, that respect the morphology of the training shapes.

In this work, we depart from existing approaches and propose a new method that leverages a statistical model of shape, built in a strongly supervised way, akin to that of ASM and AAM, while retaining the advantages of GANs. We do so by imposing a shape prior on the output of the generator. We enforce the corresponding geometry on the object outputted by the generator using a differentiable geometric function that depends on the shape prior. Our method does not require complex architectures and can be used to augment any existing GAN architecture. 
 
\section{Geometry-Aware GAN}
\label{sec:method}
In GAGAN, we disentangle the input random noise vector $\myvector{z}$ to enforce a geometric prior and learn a meaningful latent representation. To do so, we model the geometry of objects using a collection of fiducial points. The set of all fiducial points of a sample composes its shape. Using the set of these shapes on the training set, we first build a statistical shape model capable of compactly representing them as a set of normal distributed variables. We enforce that geometry by conditioning the output of the generator on shape parameter representation of the object. The discriminator, instead of being fed the output of the generator, sees the images mapped onto the canonical coordinate frame by a differentiable geometric transformation (\emph{motion model}). 

\begin{figure}[ht!]
\begin{center}
   \includegraphics[width=1\linewidth]{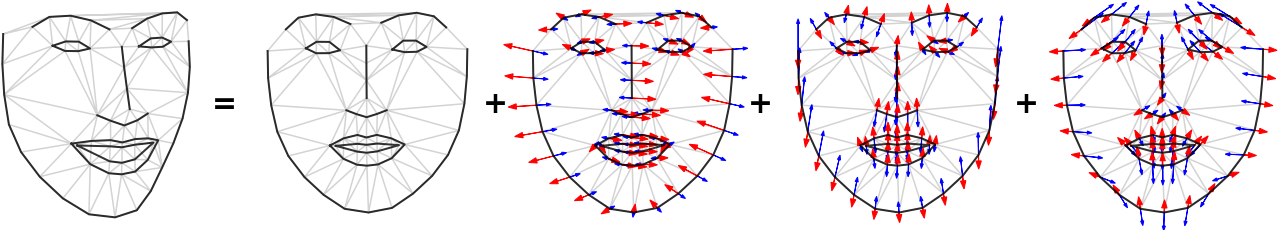}
\end{center}
   \caption{\textbf{Illustration of the statistical model of shape}. An arbitrary shape can be expressed as a canonical shape plus a linear combination of shape eigenvectors. These components can be further interpreted as modeling pose (components 1 and 2), smile/expression (component 3), etc.}
\label{fig:shape_space}
\end{figure}
\paragraph{Building the shape model}
Each shape, composed of \(m\) fiducial points is represented by a vector of size \(2m\) of their 2D coordinates \(\myvector{x}_1, \myvector{y}_1, \myvector{x}_2, \myvector{y}_2, \cdots, \myvector{x}_m, \myvector{y}_m\).
First, similarities -- that is, translation, rotation and scaling-- are removed from these using Generalised Procrustes Analysis \cite{cootes1995active}. Principal Component Analysis is then applied to the similarity free shapes to obtain the mean shape \(\myvector{s}_0\) and a set of eigenvectors (the principal components) associated with the eigenvalues . The first \(n - 4\) eigenvectors associated with the largest eigenvalues \(\lambda_1, \cdots, \lambda_n\) are kept and these compose the shape space. However, since this model was obtained on similarity free-shapes it is unable to model translation, rotation and scaling. We therefore mathematically build \(4\) additional components to model these similarities and append them to the model before re-orthonormalising the whole set of vectors \cite{matthews2004active}.
By stacking the set of all \(n\) components as the columns of a matrix \(\mymatrix{S}\) of size \((2m, n)\), we obtain the shape model.
A shape \(\myvector{s}\) can then be expressed as:
\begin{equation} \label{shape_model}
    \myvector{s} = \myvector{s}_0 + \mymatrix{S}\myvector{p}, 
\end{equation}

We define \(\phi\) the mapping from the shape space to the parameter space:
\begin{align*}
  \phi \colon \myR^{2m} &\to \myR^{n}\\
  \myvector{s} &\mapsto \mymatrix{S}\myT(\myvector{s} -  \myvector{s}_0) = \myvector{p} 
\end{align*}

\begin{figure}[t]
\begin{center}
   \includegraphics[width=0.8\linewidth]{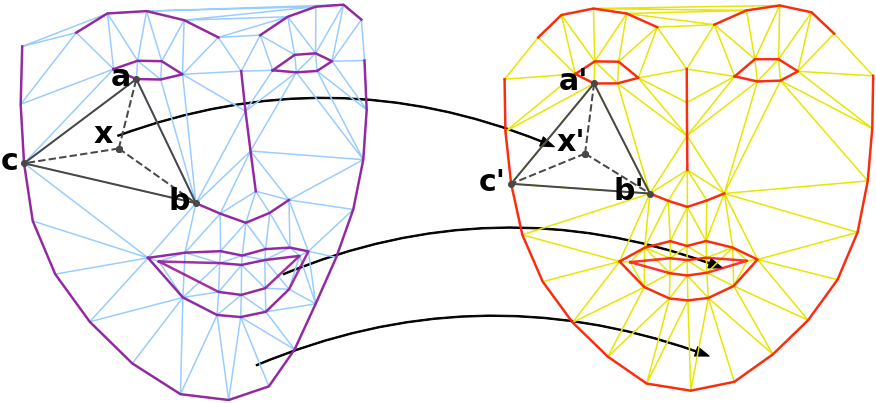}
\end{center}
   \caption{\textbf{Illustration of the piecewise affine warping from an arbitrary shape (left) onto the canonical shape (right)}. After the shapes have been triangulated, the points inside each of the simplices of the source shape are mapped to the corresponding simple in the target shape. Specifically, a point \(x\) is expressed in barycentric coordinates as a function of the vertices of the simplex it lays in. Using these barycentric coordinates, it is mapped onto \(x'\) in the target simplex.}
\label{fig:warp}
\end{figure}

This transformation is invertible, and its inverse, \(\phi^{-1}\) is given by \(\phi^{-1} \colon \myvector{p} \mapsto \myvector{s}_0 + \mymatrix{S}\mymatrix{S}\myT(\myvector{s} -  \myvector{s}_0) 
\).

We can interpret our model from a probabilistic standpoint \cite{davies2008statistical}, where the shape parameters \(\myvector{p}_1, \cdots, \myvector{p}_n\) are independent Gaussian variable with variance \(\lambda_1, \cdots, \lambda_n\) and zero mean. By using the normalised shape parameters \(\frac{\myvector{p}_1}{\sqrt{\lambda_1}}, \cdots, \frac{\myvector{p}_n}{\sqrt{\lambda_n}}\), we enforce that they be independent and normal distributed, suitable as input to our generator. This also gives us a criteria to assess how realistic a shape is using the sum of its normalised parameters \( \sum_{k=1}^n \frac{\myvector{p}_k}{\sqrt{\lambda_k}} \sim \chi^2\), which follows a Chi squared distribution \cite{davies2008statistical}.

\begin{figure*}[ht!]
\includegraphics[width=\textwidth]{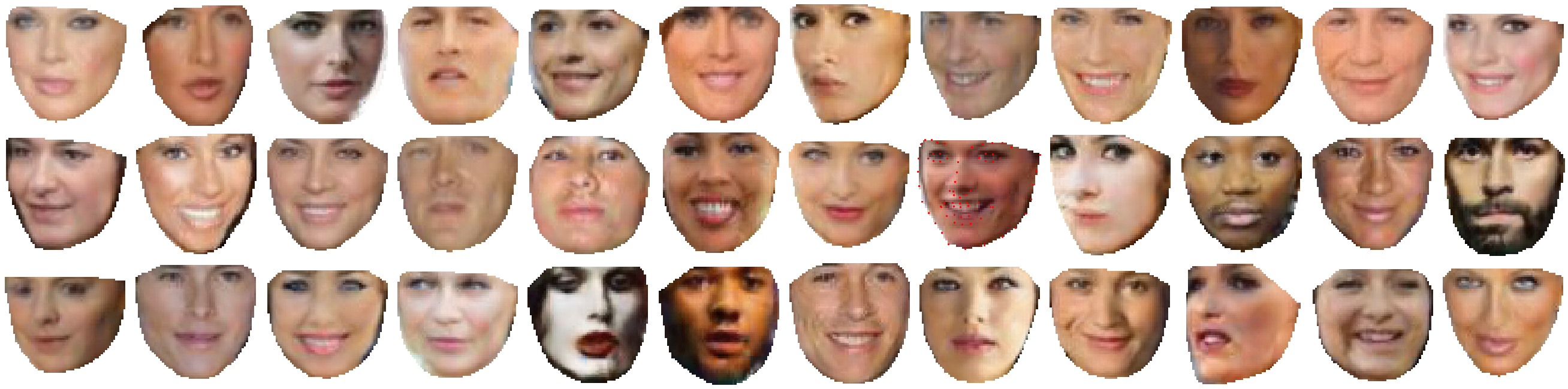}
\caption{Random 64x64 samples from GAGAN (ours).}
\label{fig:samples}
\end{figure*}

\paragraph{Enforcing the geometric prior}

To constrain the output of the generator to correctly respect the geometric prior, we propose the use of the differentiable geometric function. Specifically, the discriminator never directly sees the output of the generator. Instead, we leverage a motion model that, given an image and a corresponding set of landmarks, maps the image onto the canonical coordinate frame. The only constraint on that motion model is that it has to be differentiable. We then backpropagate from the discriminator to the generator, through that transformation. 

In this work, we use a piecewise affine warping as the motion model. The piecewise affine warping works by mapping the pixels from a source shape onto the target shape. In this work, we employ the canonical shape. This is done by first triangulating both shapes, typically as a Delaunay triangulation. The points inside each simplex of the source shape are then mapped to the corresponding triangle in the target shape, using its barycentric coordinates in terms of the vertices of that simplex, and the corresponding value is decided using the nearest neighbor or interpolation. This process is illustrated in Figure~\ref{fig:warp}.

\paragraph{GAGAN}
We consider our input as a pair of N images \(\mymatrix{I} \in \mathbb{R}^{N \times h \times w }\) and their associated shapes (or set of fiducial points) \(\myvector{s} \in \mathbb{N}^{N \times k \times 2}\), where $h$ and $w$ represent height and width of a given image, and $k$ denotes the number of fiducial points. From each shape \(\myvector{s}^{(i)}, \; i=1,\ldots,N\), we generate \(K\) perturbed version: \(\myvector{\hat{s}}^{(i)} = (\myvector{\hat{s}}^{(i)}_1, \ldots,  \myvector{\hat{s}}^{(i)}_K)\). We denote \(\myvector{\hat{p}}^{(i)} = (\myvector{\hat{p}}^{(i)}_1, \cdots,  \myvector{\hat{p}}^{(i)}_K)\) their projection onto the normalised shape space, obtained by $\myvector{\hat{p}}^{(i)}_j = \Phi(\myvector{s}^{(i)}_j), \; j=1, \cdots, K$. We model $\myvector{\hat{p}}^{(i)}_j \sim \mathcal{N}(0, 1)$ as a set of structured latent variables which represents the geometric shape of the output objects.
For simplicity, we may assume a factored distribution, given by 
\(
P(\myvector{\hat{p}}^{(i)}_{1}, \ldots, \myvector{\hat{p}}^{(i)}_{n}) = \prod_j P(\myvector{\hat{p}}^{(i)}_{j}), \; i = 1,\cdots,N, \; j=1,\cdots,n\).

We now propose a method for discovering these latent variables in a supervised way: we provide the generator network $G$ with both a the latent code vector $\myvector{\hat p}^{(i)}_j$ and an associated noise vector $\myvector{c}^{(i)}_j$, so the form of the generator becomes $G(\myvector{c}^{(i)}_j, \myvector{\hat p}^{(i)}_j)$. 
However, in standard GAN and given a large latent space, the generator is able to ignore the additional latent code $\myvector{p}^{(i)}_j$ by finding a solution satisfying $P_G(\myvector{x}^{(i)}|\myvector{p}^{(i)}_j) = P_G(\myvector{x}^{(i)})$. To cope with the problem of trivial latent representation, we propose to employ a differentiable geometric transformation $\mathcal{W}$, also called motion model, that maps the appearance from a generated image to a canonical reference frame. In this work, we employ a piecewise affine warping and map onto the mean shape \(\myvector{s}_0\)). The discriminator only sees fake and real samples after they have been mapped onto the mean shape. Discriminating between real and fake is then equivalent to jointly assessing the quality of the appearance produced as well as the accuracy of the shape parameters on the generated geometric object. The use of a piecewise affine warping has an intuitive interpretation: the better the generator follows the given geometric shape, the better the result when warping to the mean shape. For ease of notation, we denote $\myvector{\hat{z}}^{(i)}$ the latent variable concatenating $\myvector{\hat{p}}^{(i)}$ and $\myvector{c}^{(i)}$, $\myvector{\hat{z}}^{(i)} = \bm{(}\myvector{\hat{p}}^{(i)}, \myvector{c}^{(i)}\bm{)}$.

We propose to solve the following affine-warping-regularized minimax game:

\begin{align}
\begin{split}
\min_G \max_D V(D, &G) =
\myexpectation{
	\myvector{I}, \myvector{s} \sim P_{data}}{
	\log D \bm{\big(\,}\mathcal{W}(\myvector{I}, \myvector{s})\bm{\big)\,}}
\\
&
+ \myexpectation{
	\myvector{\tilde{z}} \sim \mathcal{N}(0, 1)}{
	\log\mylp 1 - 
    	D\bm{\big(}
        	\mathcal{W}\bm(G(\myvector{\tilde{z}}), \myvector{\hat{s}}\bm)
          \bm{\big)}
        \myrp}
\end{split} \label{eq:gagan}
\end{align}

\begin{figure*}
\centering
\begin{subfigure}[b]{0.45\textwidth}
\includegraphics[width=1\linewidth]{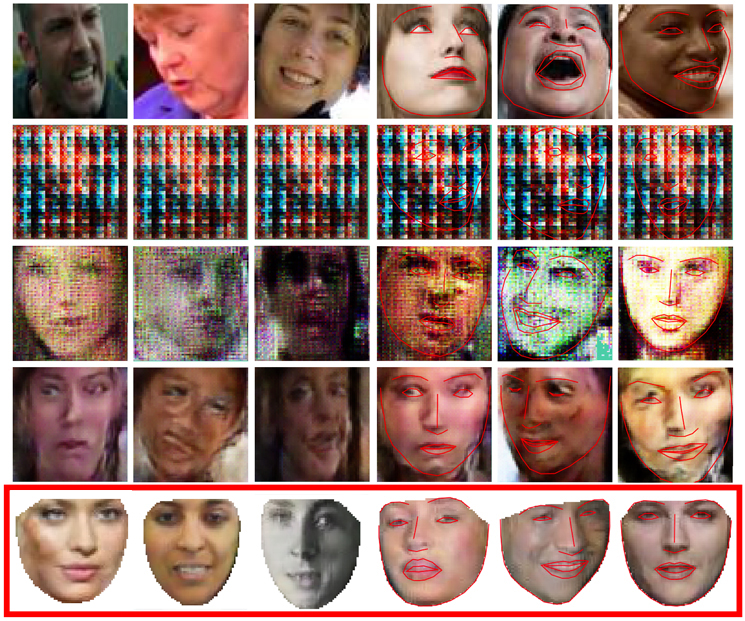}
\caption{GAGAN small set}
\label{fig:comparison:300vw}
\end{subfigure}
\hfil
\begin{subfigure}[b]{0.45\textwidth}
\includegraphics[width=1\linewidth]{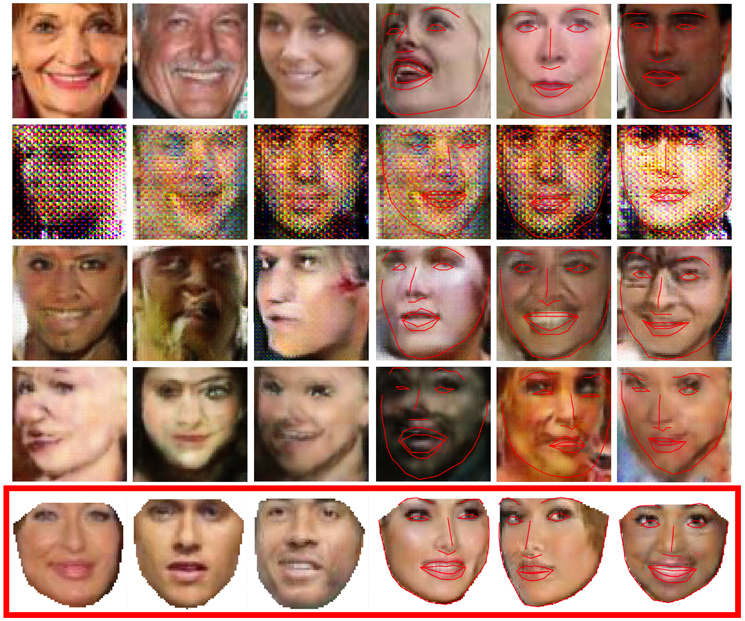}
\caption{CelebA}
\label{fig:comparison:celebA}
\end{subfigure}
\caption{\textbf{Comparison between samples of faces generated by the baseline models and our model GAGAN for the GAGAN-small set (left column) and celebA (right column)}. The first row shows some real images. The following rows presents results obtained with our baseline models: row (2): Shape-CGAN, row (3): P-CGAN and row (4): Heatmap-CGAN. The last row present some images generated by our proposed GAGAN architecture. The first three columns show generated samples solely, while we visualize the shape prior, overlaid on the generated images, in the last three columns.}
\label{fig:comparison}
\end{figure*}

\paragraph{Local appearance preservation}
The generative model of shape provides us rich information about the images being generated. In particular, it is desirable for the appearance of a face to be dependent on the set of fiducial points that compose it (i.e. an infant's face has a different shape and appearance from that of an adult male or female face). However, we also know that certain transformations should preserve appearance and identity. For instance, differences in head pose should ideally not affect appearance. 

To enforce this, rather than directly feeding the training shapes to the generator, we feed it several appearance-preserving variations of each shape, and ensure that the resulting samples have similar appearance. Specifically, for each sample, we generate several variants by mirroring it, projecting it into the normalised shape space, adding random noise sampled from a Gaussian distribution there, and using these perturbed shapes as input. Since the outputs should look different (as they have different poses for instance), we cannot directly compare them. However, the geometric transformation projects these onto a canonical coordinate frame where they can be compared, allowing us to add a loss to account for these local appearance preservations.

Formally, we mirror the images and denote the corresponding mirrored shape and shape parameter are denoted by $\myvector{\hat{s}}^{(i)}_{M}$ and $\myvector{p}^{(i)}_{jM}$. The mirrored, normalised shape parameters $\myvector{\hat{p}}_M$ are then used to build the latent space $\myvector{\hat{z}}_M \sim \mathcal{N}(0, 1)$. For simplicity, we define perturbed shapes
$\myvector{\tilde{s}} = (\myvector{\hat{s}}, \myvector{\hat{s}}_M)$,
normalised parameters
$\myvector{\tilde{p}} = (\myvector{p}, \myvector{p}_M)$ 
and latent vectors $\myvector{\tilde{z}} = (\myvector{\hat{z}}, \myvector{\hat{z}}_M)$ that share a common noise vector \(\myvector{c}\).
Finally, we define the mirroring function $m(\cdot)$, that flips every image or shape horizontally. The local appearance preservation loss (LAP) is then defined as:
\begin{align}
\begin{split}
LAP = & \mynorm{
\mathcal{W}\mylp G(\myvector{\hat{z}}),\myvector{\hat{s}}\myrp
- \mathcal{W}\mylp m\bm{\big(}G(\myvector{\hat{z}}_M)\bm{\big)}, m\left(\myvector{\hat{s}}_M\right)\myrp
} + \\
&  \mynorm{
\mathcal{W} \mylp m\bm{\big(}G(\myvector{\hat{z}})\bm{\big)}, m(\myvector{\hat{s}}) \myrp -
\mathcal{W} \mylp G(\myvector{\hat{z}}_M), \myvector{\hat{s}}_M \myrp
}
\end{split}
\end{align}

Adding the local appearance preservation to the minimax optimization value function, we get the final objective:
\begin{align}
\begin{split}
\min_G \max_D V(D, &G) =
\myexpectation{
	\myvector{I}, \myvector{s} \sim P_{data}}{
	\log D \bm{\big(\,}\mathcal{W}(\myvector{I}, \myvector{s})\bm{\big)\,}}
\\
&
+ \myexpectation{
	\myvector{\tilde{z}} \sim \mathcal{N}(0, 1)}{
	\log\mylp 1 - 
    	D\bm{\big(}
        	\mathcal{W}\bm(G(\myvector{\tilde{z}}), \myvector{\tilde{s}}\bm)
          \bm{\big)}
        \myrp}\\
& + \lambda \cdot LAP
\end{split}
\end{align}
A visual overview of the method can be found in Figure~\ref{fig:GAGAN} and Figure~\ref{fig:samples} presents samples generated with GAGAN.

\section{Experimental results}
\label{sec:experiment}
In this section, we introduce the experimental setting and 
demonstrate the performance of the GAGAN quantitatively and qualitatively on 
what is arguably the most popular application for GANs, namely 
face generation. 
Experimental results indicate that the proposed method outperforms existing architectures while respecting the geometry of the faces.

\subsection{Experimental setting}

\paragraph{Datasets}
To train our method, we used widely established databases for facial landmarks estimation, namely Helen \cite{Helen}, LFPW \cite{LFPW}, AFW \cite{AFW} and iBUG \cite{sagonas2013semi}. In all cases we used \(68\) landmarks, in the Multi-Pie configuration \cite{gross2010multi} as annotated for the \(300\)-W challenge \cite{sagonas2013faces,sagonas2013semi}. We also used the test set of the \(300-W\) challenge \cite{300W} and sampled frames from the video of the \(300\)-VW challenge \cite{300VW}, as well as the videos of the AFEW-VA dataset \cite{afewva}. We coin the set of all these images and shapes the \emph{GAGAN-small set}.
To allow for comparison with other traditional GAN methods, we also used the CelebA dataset \cite{liu2015faceattributes}, which contains \(202,599\) images of celebrities. Finally, to demonstrate the versatility of the method, we apply it to the cat faces dataset introduced in \cite{cats_iccv,cats_ijcv}

\paragraph{Pre-processing}
All images where processed in the following way: first the shape in the image was rescaled to a size of \(60 \times 60\). The corresponding image was resized using the same factors and then cropped into a size of \(64 \times 64\) so that the shape is in the center with a margin of \(2\) pixels on all sides. Since the celebA dataset is only annotated for \(5\) fiducial points, we use the recent deep learning based face alignment method introduced in \cite{bulat2017far} to detect these. This method has been shown to provide remarkable accuracy, often superior to that of humans annotators \cite{bulat2017far}.

\begin{figure*}
\centering
\begin{subfigure}[b]{0.45\textwidth}
  \includegraphics[width=1\linewidth]{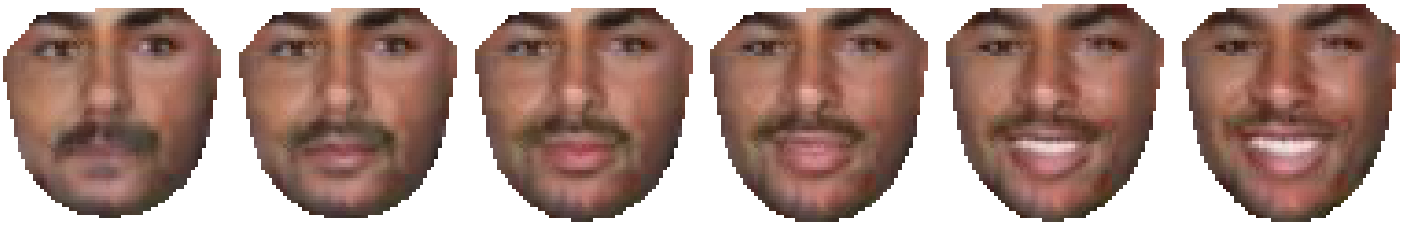}\\%
  \includegraphics[width=1\linewidth]{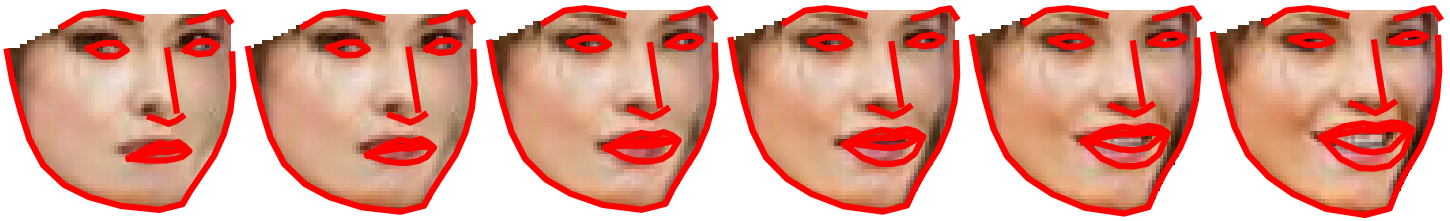}\\%
  \includegraphics[width=1\linewidth]{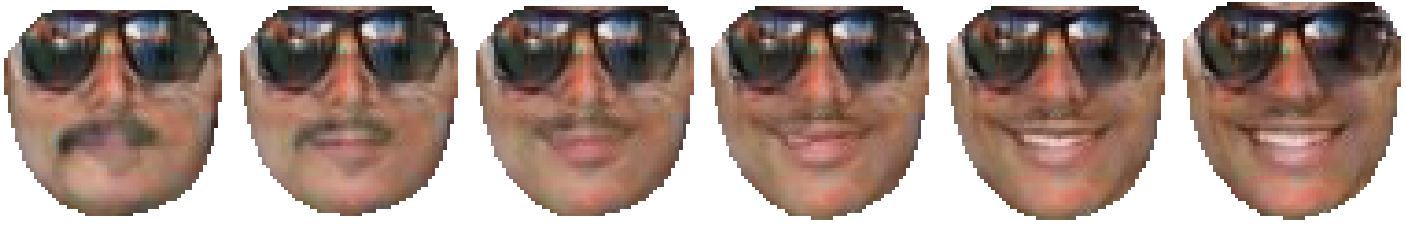}\\%
  \includegraphics[width=1\linewidth]{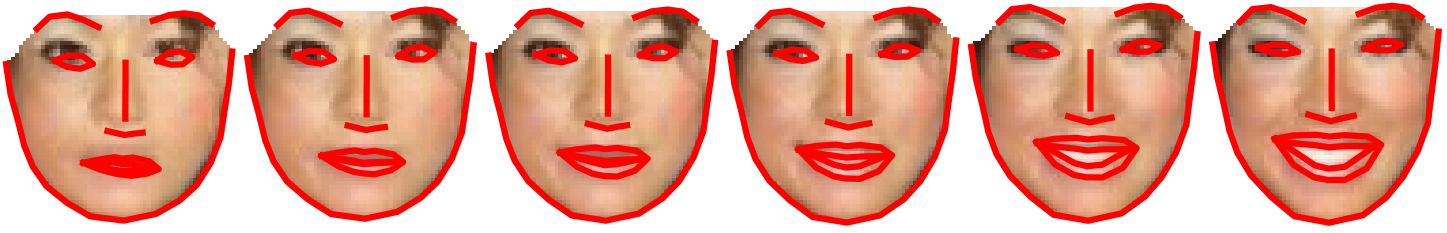}\\%
  \includegraphics[width=1\linewidth]{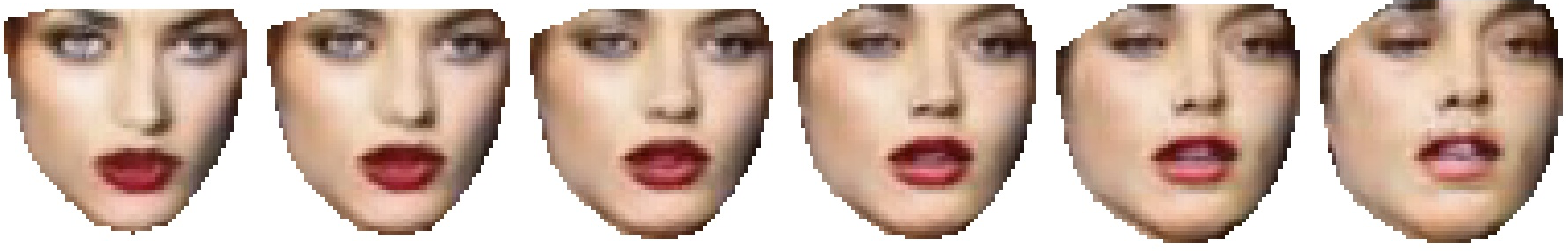}
  \caption{Varying the shape parameters}
  \label{fig:interp_same_noise}
\end{subfigure}
\hfil\hfil
\begin{subfigure}[b]{0.45\textwidth}
  \includegraphics[width=1\linewidth]{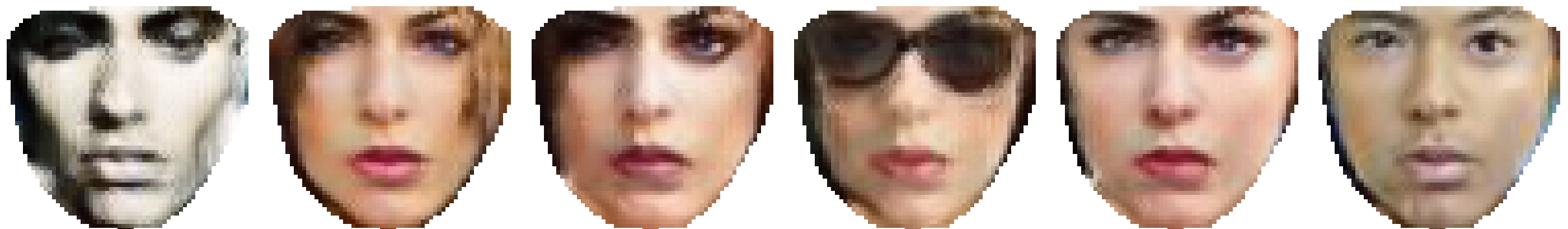}\\%
  \includegraphics[width=1\linewidth]{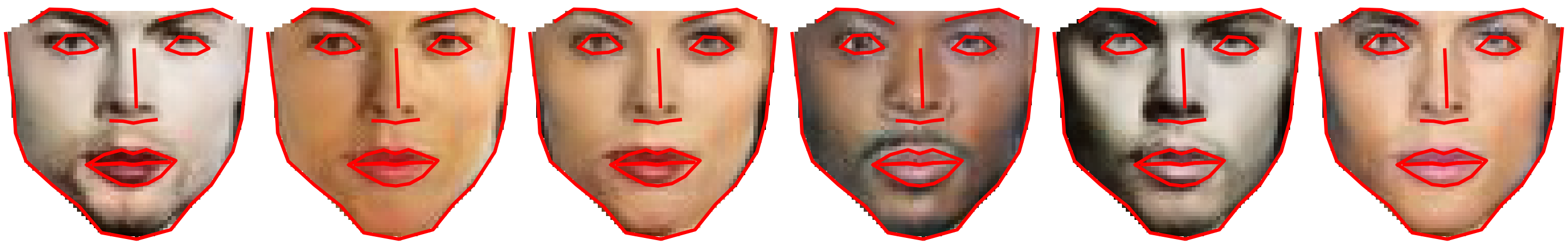}\\%
  \includegraphics[width=1\linewidth]{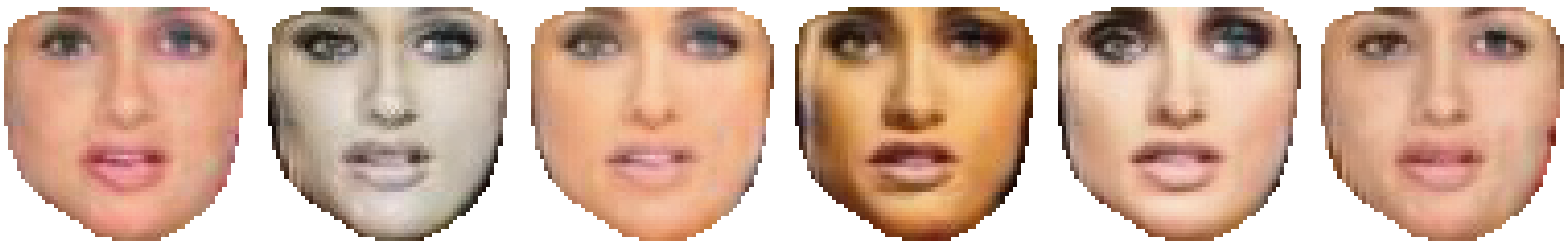}\\%
  \includegraphics[width=1\linewidth]{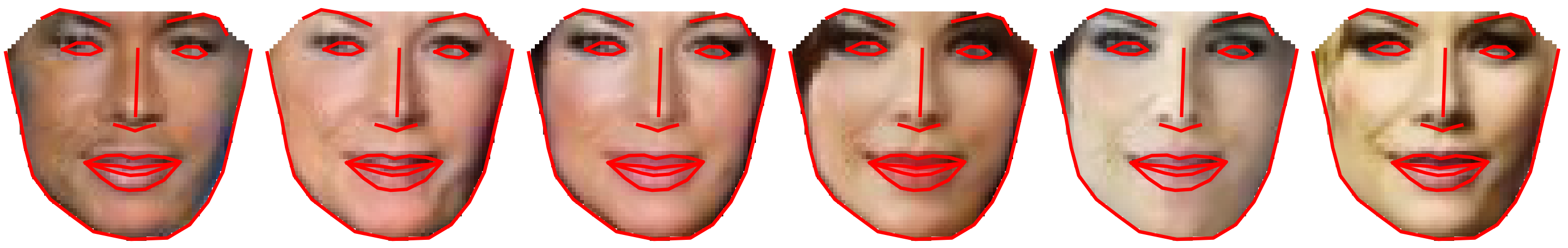}\\%
  \includegraphics[width=1\linewidth]{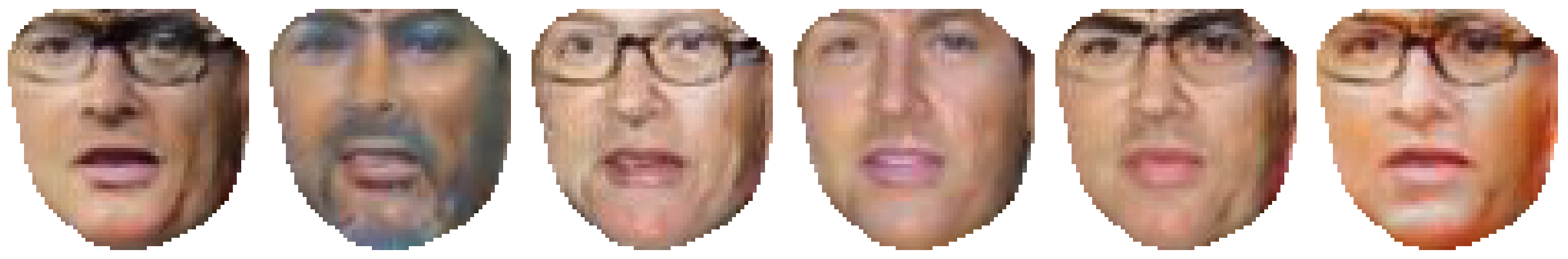}
  \caption{Varying the appearance parameters}
  \label{fig:interp_same_shape}
\end{subfigure}
\caption{\textbf{Images obtained by varying only the parameters from the shape prior (left) or non geometric prior (right)}.
In (\ref{fig:interp_same_noise}), we vary only a few shape parameters, while keeping all others fixed. The components of the statistical shape space can easily be interpreted in term of pose, morphology, smile, etc.
Conversely, by varying only the non geometric latent vector \(\myvector{c}\) and keeping the shape parameters fixed (\ref{fig:interp_same_shape}), we can control appearance and generate more diverse images.}
\label{fig:interpolations}
\end{figure*}

\paragraph{Implementation and training details} We used a standard DC-GAN architectures as defined in \cite{radford2015unsupervised}, with input images of size \(64 \times 64\). The latent vector \(\myvector{z}\) of the generator has size \(100\) and is composed of the \(50\) normalised shape parameters concatenated with i.i.d. random noise sampled from a normal distribution. We trained our model using Adam \cite{diederik2014adam}, with a learning rate of 0.0002 for the discriminator and a learning rate of 0.001 for the generator. Model collapse has been observed with high learning rates. Reducing the learning rate was sufficient to avoid this issue. We used $\lambda$ in range $[0.0, 1.0, 2.5, 5.0]$. We found $2.5$ to be the best regularization factor in terms of quality of generated images. All experiments were ran on a single GPU on Amazon Web Services, with an NVIDIA VOLTA GPU.

\paragraph{Baseline models} For our baselines, we used 3 models, which the same architecture as DCGAN \cite{radford2015unsupervised}:
\begin{description}
\item{\emph{Shape-CGAN}} is a Conditional GAN (CGAN) \cite{mirza2014conditional}, modified to generate images conditioned on shapes \(\myvector{s}\) by channel-wise concatenation.
\item{\emph{P-CGAN}} is a CGAN conditioned on the normalised shape parameters \(\myvector{p}\) (as used by GAGAN and introduced previously) by channel-wise concatenation.
\item{\emph{Heatmap-CGAN}} is a novel model, based on a CGAN conditioned on shapes by heatmap concatenation. First a heatmap with value \(1\) at the expected position of landmarks, and \(0\) everywhere else is created. This is then used as an additional channel and concatenated to the image passed on to the discriminator. For the generator, the shapes are flattened and concatenated to the latent vector $\myvector{z}$ obtained from our statistical shape model.
\end{description}

\subsection{Qualitative results}
Figure \ref{fig:samples} shows some representative samples drawn from $\myvector{z}$ at a resolutions of 64 x 64. We observe realistic images that closely follow the imposed shape prior, for a wide range of poses, expression, gender and illumination. Though we observed fewer older people, the proportion between men and women sampled appears to be balanced. Interestingly, the model was able to generate accessories, such as glasses, during sampling.

We also compared the quality of images generated by GAGAN and our baseline models (Fig.~\ref{fig:comparison}), on the GAGAN small set and CelebA datasets. When trained on GAGAN small set, (Fig.~\ref{fig:comparison:300vw}), Shape-CGAN fails to generate any meaningful image. P-CGAN, on the other hand, generates images of faces that respect the shape parameters, validating the use of such a representation. However, the generated images are highly pixelated and textures are rudimentary. Heatmap-GAN correctly generates faces according to the shapes and the textures are more realistic than P-CGAN, but the geometry is distorted. Our model, GAGAN, generates the most realistic images among all models and accurately follows the shape prior. On CelebA, generation is better for all models, including ours (Fig.~\ref{fig:comparison:celebA}). As observed on the small set, the baseline models can generate meaningful images that approximately follow the shape prior, but inferior to that of GAGAN, either low quality (Shape-CGAN), or highly distorted (P-CGAN, Heatmap-CGAN).
The difference in performance between the two datasets can be explained by their size, CelebA being about ten times as large as GAGAN small set. As is known, deep learning methods, including GANs, typically work best with large datasets. 

\subsection{Quantitative results}
The facial landmark detector introduced in \cite{bulat2017far} is reported to
detect fiducial points with an accuracy in most cases higher than that of human annotators.
Our model takes as input a shape prior and generates an image that respects that prior. Therefore, we propose to assess the quality of the results by running the landmark detector on the produced images and measuring the distance between the shape prior and the actual detected shape. We directly run the detector on \(10,000\) images created by the generator of our GAGAN, heatmap-CGAN and P-CGAN, all trained on CelebA. 

\begin{figure}[ht!]
\centering
\includegraphics[width=1\linewidth]{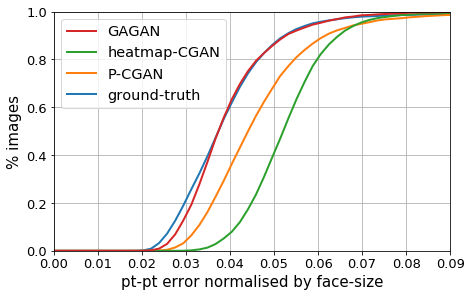}
\label{DPM_LFPW_3}
\caption{
\textbf{Cumulative Error Distribution}.
We plot the error between the landmarks estimated by the detector and those used as prior to generate the images, for GAGAN (red), heatmap-CGAN (green) and P-CGAN (orange), all trained on CelebA.
As a baseline, we evaluate the performance of the landmark detector on our GAGAN-small set (\emph{ground-truth}, blue).
}
\label{fig:quantitative}
\end{figure}

Performance is evaluated in terms of the well-established normalised point-to-point error (\emph{pt-pt-error}), as introduced in \cite{AFW} and defined as the RMS error, normalised by the face-size.
Following \cite{kossaifi_newton_aam,tzimiropoulos2016fast,AFW,kossaifi_bidirectional_aam,tzimiropoulos2017fast,kossaifi2017fast},
we produced the cumulative error distribution (CED) curve, Fig.~\ref{fig:quantitative}. It depicts, for each value on the x-axis, the percentage of images for which the point-to-point error is lower than this value. As a baseline (\emph{ground-truth}), we run the facial landmark detector on our challenging GAGAN-small-set, and compute the errors with the annotations provided with the data. 

As can be observed, the images generated by GAGAN accurately follow the given geometric prior used for generation, with an accuracy similar to that of the landmark detector. While heatmap-CGAN and P-CGAN also generate images that follow the prior, they do so with a significantly lower accuracy, which might also be due to the lesser quality of the images generated.

\subsection{Generality of the model}
To demonstrate the versatility of the model, we apply it to the generation of cats faces, using the dataset introduced in \cite{cats_iccv,cats_ijcv}. Specifically, we used \(348\) images of cats, for which \(48\) facial landmarks were manually annotated \cite{cats_iccv}, including the ears and boundaries of the face. We build a statistical shape space as previously done for human faces and condition GAGAN on the resulting shape parameters. 

\begin{figure}[ht!]
\centering
\includegraphics[width=0.19\linewidth]{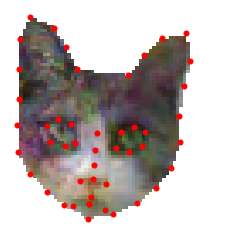}
\includegraphics[width=0.19\linewidth]{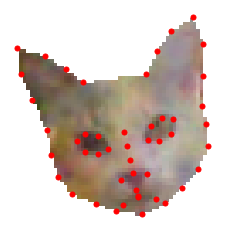}
\includegraphics[width=0.19\linewidth]{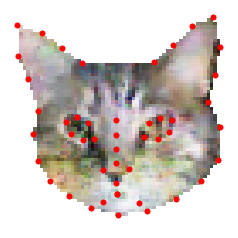}
\includegraphics[width=0.19\linewidth]{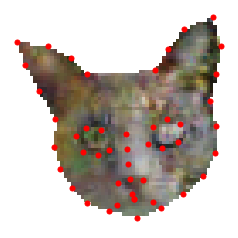}
\includegraphics[width=0.19\linewidth]{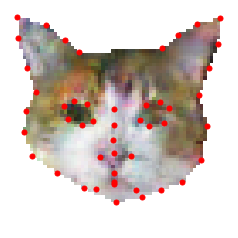}\\
\includegraphics[width=0.19\linewidth]{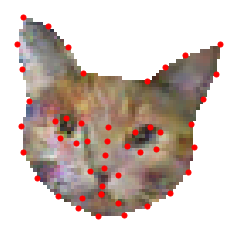}
\includegraphics[width=0.19\linewidth]{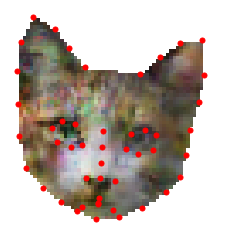}
\includegraphics[width=0.19\linewidth]{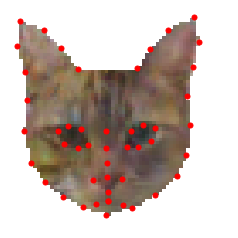}
\includegraphics[width=0.19\linewidth]{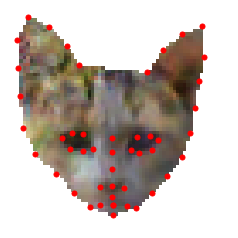}
\includegraphics[width=0.19\linewidth]{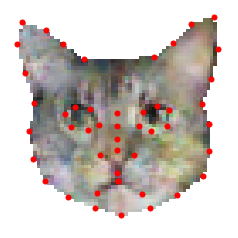}
\caption{
Samples generated by our model trained on the cats dataset, overlaid with the geometric prior used for generation (red points).
}
\label{fig:cats}
\end{figure}

We present some examples of generated images, along with the geometrical prior used for generation in Figure~\ref{fig:cats}.

\section{Conclusion and future work}
\label{sec:conclusion}
We introduced GAGAN, a novel method that can be used to augment any existing GAN architecture to incorporate geometric information. Our generator samples from the probability distribution of a statistical shape model and generates faces that respect the induced geometry. This is enforced by an implicit connection from the shape parameters fed to the generator to a differentiable geometric transform applied to its output. The discriminator, being trained only on images normalised to a canonical image coordinates is able to not only discriminate on whether the produced fakes are realistic but also on whether they respect the geometry.
As a result, our model is the first one, to wit, able to produce realistic images conditioned on an input shape.
Going forward, we are currently working on extending our method in several ways by, i) applying it to the generation of larger images, ii) exploring more complex geometric transformations that have the potential to alleviate the deformations induced by the piecewise-affine warping and iii) augmenting traditional CNN architectures with our method for facial landmark detection. 

\section{Acknowledgements}
The authors would like to thank Amazon Web Services for providing the computational resources to run the experiments of these paper.
The work of Jean Kossaifi, Yannis Panagakis and Maja Pantic has been funded by the
European Community Horizon 2020 [H2020/2014-2020] under grant
agreement no. 645094 (SEWA).
The work of Linh Tran has been funded by the European Community
Horizon 2020 under grant agreement no. 688835 (DE-ENIGMA).

{\small
\bibliographystyle{ieee}

}

\end{document}